\documentclass[11pt]{article}
\usepackage[T1]{fontenc}   
\usepackage[margin=1in]{geometry}
\usepackage{graphicx}
\usepackage{booktabs}
\usepackage{amsmath,amssymb}
\usepackage{xcolor}
\usepackage{enumitem}
\usepackage[numbers,sort&compress]{natbib}
\usepackage[hidelinks]{hyperref}
\usepackage{xurl}   
\usepackage{tikz}
\usetikzlibrary{arrows.meta,positioning}
\usepackage[ruled,vlined]{algorithm2e}
\usepackage{placeins}
\usepackage{caption}
\graphicspath{{figures/}}

\newcommand{\Dij}{D_{ij}}
\newcommand{\code}[1]{\texttt{#1}}

\title{\textbf{GNN-Guided Graph Coarsening and Adaptive QUBO Penalties for the
Capacitated Vehicle Routing Problem with Time Windows on a Quantum Annealer}}
\author{
  Youssef Kamel Rezk$^{1}$ \qquad Pawe\l{} Gora$^{2,3}$ \\[6pt]
  {\small $^{1}$Alamein International University, Alamein, Egypt}\\[1pt]
  {\small $^{2}$Jagiellonian University, Krak\'ow, Poland \qquad
   $^{3}$Fundacja Quantum AI, Warsaw, Poland}\\[3pt]
  {\small \texttt{youssefkamelkamel67@gmail.com} \qquad \texttt{pawel.gora@qaif.org}}
}
\date{\today}

\begin{document}
\maketitle

\begin{abstract}

Graph coarsening reduces the large Quadratic Unconstrained Binary Optimization (QUBO) formulations arising when vehicle-routing problems are solved by quantum annealing. Nearby customers with compatible time windows are merged into super-nodes, the reduced problem is solved, and the solution is expanded to the original graph. For the Capacitated Vehicle Routing Problem with Time Windows (CVRPTW), however, existing coarsening heuristics require family-specific tuning and remain unreliable on random instances. We address these limitations on the Solomon benchmark using simulated annealing and a D-Wave Advantage2 processor.

First, we introduce adaptive penalty calibration. Uniform penalty scaling has little effect, whereas controlling the internal coefficient range—dominated by non-binding capacity log-slack terms on most small instances—substantially improves raw samples. Removing non-binding constraints, normalising binding ones, and scaling the remaining penalties reduces mean raw constraint violations from 33.0 to 0.06 at the same solver budget (paired Wilcoxon test, $p=3.7\times10^{-11}$, $n=56$). A variable-count-preserving control attributes this gain to conditioning rather than problem size.

Second, we replace the hand-tuned merge score with a graph neural network (GNN) using one configuration across all families. At $N=10$, it achieves 100\% feasibility across all Solomon families, including R-type (100\% vs. 80\% for the tuned heuristic). Across $N\in{10,\dots,100}$, feasibility is 83\% vs. 69\%, with the GNN better or tied on 85/90 instance-size pairs. The difference is significant at $N=80,100$ ($p=0.002$; 25/25 pairs), while the QUBO remains approximately 5–6× smaller.

Finally, hardware experiments reproduce the conditioning effect at fixed logical variable count: feasible samples increase from 0.02\% to 39\% across 13 instances. These results concern QUBO conditioning, feasibility, and tractability; for end-to-end solution cost, classical repair with local search remains a reference bound that the pipeline matches but does not surpass.

\end{abstract}

\section{Introduction}
The Vehicle Routing Problem (VRP) and its time-windowed variant (CVRPTW) are
widely studied combinatorial-optimization problems with direct applications in
transportation and logistics. Quantum annealers and their classical simulators
minimise QUBO energy functions, motivating several QUBO formulations of routing
problems \citep{borowski2020,nalecz2025,ozyilmaz2025}. Their principal limitation is
size: a direct VRP encoding grows rapidly and only small instances fit on current
hardware. \emph{Graph coarsening} reduces this burden by repeatedly merging
compatible customers into super-nodes, solving the reduced problem, and expanding
the solution to the original graph \citep{nalecz2025,ozyilmaz2025}.

The method of \citet{ozyilmaz2025} extends coarsening to time windows through a
spatio-temporal merge metric. Its performance depends on the instance family:
clustered (C) instances remain feasible, whereas random (R) instances deteriorate
unless the coarsening hyperparameters are retuned for each family. Our work targets
this dependence on family-specific tuning.

We make four contributions, evaluated on the Solomon benchmark with a
simulated-annealing backend and on a D-Wave Advantage2 processor:
\begin{enumerate}[nosep]
  \item \textbf{Adaptive penalty calibration} (Section~\ref{sec:phase2}). We show
  that uniformly scaling all QUBO penalties leaves the sampler nearly unchanged,
  and identify the QUBO's \emph{internal dynamic range} as the effective lever.
  Dropping non-binding capacity constraints, normalising those that bind, and
  scaling the rest cuts raw constraint violations from $33.0$ to $0.06$ at equal
  budget, and an ablation attributes the gain to the capacity handling rather than
  the scaling.
  \item \textbf{GNN-guided, tuning-free coarsening} (Section~\ref{sec:phase3}). A
  GraphSAGE-based merge scorer replaces the hand-tuned heuristic and uses a single
  fixed configuration across all families, recovering $100\%$ R-type feasibility
  at $N{=}10$ without per-family tuning.
  \item \textbf{Feasibility-preserving coarsening at scale}
  (Sections~\ref{sec:phase4}--\ref{sec:scaling}). The learned coarsener keeps more
  coarsened solutions feasible than the per-family-tuned heuristic across
  $N{=}10$--$100$, with the advantage growing as the heuristic's fixed tuning fails
  to transfer and significant at the largest sizes ($p\!=\!0.002$ at $N{=}80,100$),
  while shrinking the QUBO $\sim\!5$--$6\times$ so it stays solvable at $N{=}100$.
  \item \textbf{Validation on quantum hardware} (Section~\ref{sec:qpu}). An offline
  embedding study fixes the accessible regime at $N\approx10$--$20$, and on a D-Wave
  Advantage2 processor the variable-count-preserving control separates $0.02\%$ from
  $39\%$ pre-repair feasibility, showing the conditioning mechanism survives finite
  coupler precision.
\end{enumerate}
These contributions are subject to three qualifications. A strong classical repair
step determines most of the post-repair feasibility; the pipeline does not improve
upon the OR-Tools reference, which is itself a strong metaheuristic rather than a
certified optimum; and the hardware experiments characterise the formulation, not
the comparative performance of the processor. The contribution lies in QUBO
formulation and preprocessing, and is therefore applicable to other annealing
backends.

\section{Background and related work}
\paragraph{CVRPTW.} Given a depot (node $0$) and $n$ customers with demands $d_i$,
service times $s_i$, and time windows $[e_i,l_i]$, the CVRPTW asks for a set of
vehicle routes of minimum total travel distance such that each customer is visited
once, each route's load does not exceed the vehicle capacity $Q$, and service at
each node starts within its window (arriving early incurs waiting; arriving after
$l_i$ is a violation). Travel time is Euclidean and proportional to the distance, i.e.,
$\tau_{ij}=\sqrt{(x_i-x_j)^2+(y_i-y_j)^2}$.

\paragraph{QUBO and quantum annealing.} A QUBO minimises
$E(\mathbf z)=\sum_i a_i z_i+\sum_{i<j}b_{ij}z_iz_j$ over $\mathbf z\in\{0,1\}^m$.
Quantum annealers natively minimise such energies \citep{kadowaki1998,johnson2011}.
Constrained problems are encoded by the \emph{penalty method}: constraints become
squared-violation terms weighted so that any violation costs more energy than it
can save \citep{lucas2014,glover2019}. We use a simulator, D-Wave's
\code{SimulatedAnnealingSampler} \citep{dwavesamplers}, which is the backend used
by the previous paper \citep{ozyilmaz2025}, giving unlimited, seed-reproducible runs
that isolate the algorithmic contribution from hardware noise. OR-Tools
\citep{ortools} serves as a strong classical reference.

\paragraph{Routing on annealers.} Encoding vehicle routing as a QUBO is an active
line of work. \citet{irie2019vrp} formulate a capacitated problem with time and
state variables and evaluate it on a D-Wave 2000Q, and \citet{borowski2020} define
the Full QUBO Solver (FQS) and Average Partitioning Solver (APS) route encodings we
use (Section~\ref{sec:method-qubo}). A recurring conclusion is that the position
encoding grows quickly and that sparse qubit connectivity makes minor embedding
expensive, which is the difficulty coarsening addresses.

\paragraph{Coarsening.} \citet{nalecz2025} introduced coarsening for CVRP-QUBO, and
\citet{ozyilmaz2025} extended it to time windows with the spatio-temporal merge
metric we reproduce. That line of work is the direct predecessor of this paper: it
supplies the coarsening procedure, the FQS/APS encoding, and the per-family
hyperparameter dependence we set out to remove.

\paragraph{Penalty weights.} Converting constrained problems to QUBO requires
penalty weights large enough that violating a constraint is never profitable.
Exact constructions give sufficient upper bounds, and smaller valid weights are
generally preferable because an annealer's finite coefficient precision is consumed
by whatever range the weights span \citep{ayodele2022penalty,glover2019,lucas2014}.
Our conditioning result is a statement about that range rather than about the
weights individually: we show that the ratio between the largest and smallest
coefficients, and specifically the block contributed by capacity slack, is what
determines raw solution quality.

\paragraph{Learning for combinatorial optimisation.} Learned heuristics for
combinatorial problems are surveyed by \citet{bengio2021ml4co} and, for graph neural
networks specifically, by \citet{cappart2023gnnco}. Reported gains are not uniform;
\citet{angelini2023greedy} find a GNN underperforming a greedy algorithm on maximum
independent set. Our merge scorer is
a GraphSAGE network \citep{hamilton2017} implemented in PyTorch Geometric
\citep{fey2019pyg}, and the reward-driven training is an expert-iteration loop in
the style of \citet{anthony2017,silver2018}.

\paragraph{Hardware and embedding.} A logical QUBO must be minor-embedded in the
processor's fixed connectivity graph, for which we use the heuristic of
\citet{cai2014minorminer}. Successive D-Wave topologies have increased qubit degree,
from Chimera to Pegasus to the degree-20 Zephyr graph of Advantage2
\citep{boothby2021zephyr}, and \citet{pelofske2025generations} compares the three
generations on embedded optimisation problems. Higher degree yields shorter chains
for the same logical problem, which is the property our embeddability study in
Section~\ref{sec:embed} measures directly.

\section{Method}
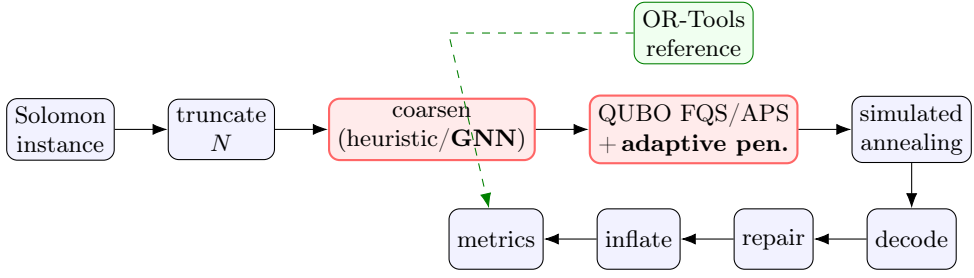
\begin{figure}[htbp]
\centering
\begin{tikzpicture}[node distance=6mm and 7mm,
  box/.style={draw, rounded corners, align=center, minimum height=8mm,
              inner sep=3pt, font=\footnotesize, fill=blue!5},
  ours/.style={box, fill=red!8, draw=red!60, thick},
  ref/.style={box, fill=green!6, draw=green!50!black},
  arr/.style={-{Latex[length=2mm]}}]
\node[box] (inst) {Solomon\\instance};
\node[box, right=of inst] (trunc) {truncate\\$N$};
\node[ours, right=of trunc] (coar) {coarsen\\(heuristic/\textbf{GNN})};
\node[ours, right=of coar] (qubo) {QUBO FQS/APS\\+\,\textbf{adaptive pen.}};
\node[box, right=of qubo] (sa) {simulated\\annealing};
\node[box, below=of sa] (dec) {decode};
\node[box, left=of dec] (rep) {repair};
\node[box, left=of rep] (inf) {inflate};
\node[box, left=of inf] (met) {metrics};
\node[ref, above=4mm of qubo] (ort) {OR-Tools\\reference};
\draw[arr] (inst)--(trunc); \draw[arr] (trunc)--(coar); \draw[arr] (coar)--(qubo);
\draw[arr] (qubo)--(sa); \draw[arr] (sa)--(dec); \draw[arr] (dec)--(rep);
\draw[arr] (rep)--(inf); \draw[arr] (inf)--(met);
\draw[arr, green!50!black, dashed] (ort) -- (met.north west |- ort) -- (met);
\end{tikzpicture}
\caption{End-to-end pipeline. Red boxes are our contributions; green is the
classical reference used to measure the gap to OR-Tools.}
\label{fig:pipeline}
\end{figure}

Figure~\ref{fig:pipeline} summarises the pipeline. Our method modifies two
components: a GNN replaces the heuristic coarsening \emph{scorer}, and adaptive
calibration replaces the static penalty scheme. OR-Tools provides the classical
cost reference.

\subsection{Spatio-temporal coarsening}
\begin{figure}[htbp]
\centering
\includegraphics[width=0.85\textwidth]{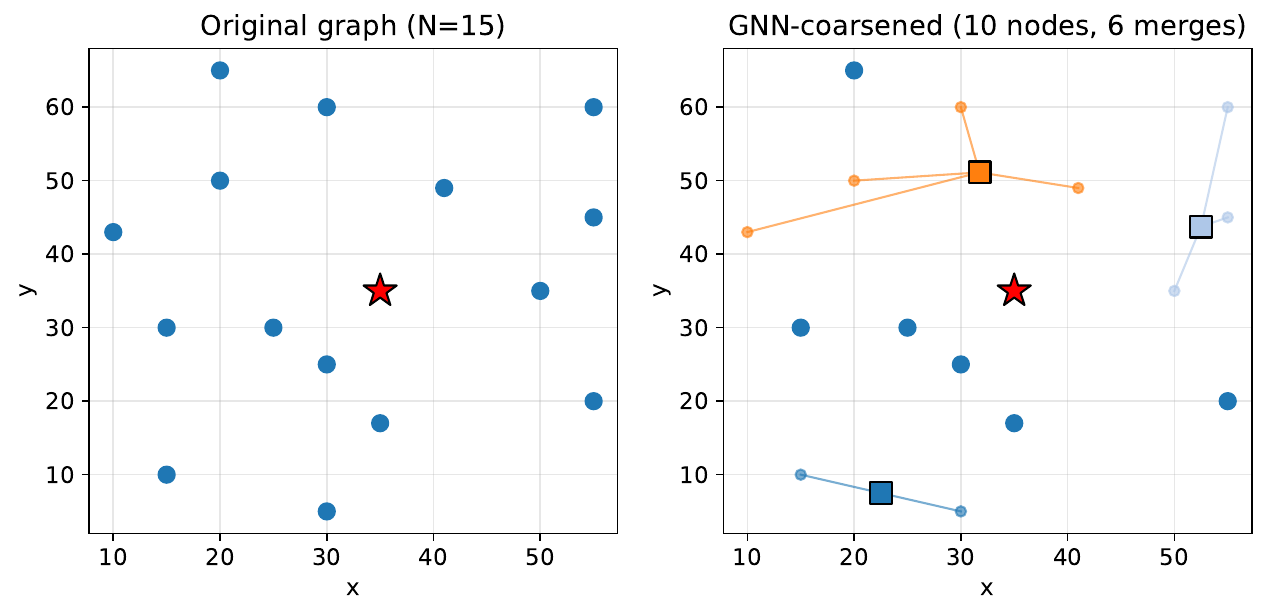}
\caption{GNN coarsening of a held-out R101 instance ($N{=}15$). Squares are
super-nodes at their midpoints; thin lines link each super-node to the original
customers it absorbs (one aggregates four via multi-level merges).}
\label{fig:coarsening}
\end{figure}

Each node has a central time $t_i=\tfrac{e_i+(l_i-s_i)}{2}$ (or $e_i$ if
$l_i-s_i<0$). The heuristic merge score (lower merges first) is
\begin{equation}
\Dij=\alpha\,\tau_{ij}+\beta\,\max\!\big(0,\;e_j-(t_i+s_i+\tau_{ij})\big),
\label{eq:dij}
\end{equation}
where $\alpha$ weights spatial proximity and $\beta$ weights temporal separation.
Coarsening proceeds in levels (Algorithm~\ref{alg:coarsen}). After all edges are
scored, a threshold $\rho$ selects the candidate fraction: edges are ranked by
$\Dij$ and $\rho$ is taken at position $\lfloor 0.1\,|E|\cdot\text{radiusCoeff}\rfloor$,
so \emph{radiusCoeff} controls how permissive a level is, with larger values
admitting more candidate pairs. A candidate pair is merged only when at least one
service order is time-feasible. The procedure then tightens the merged time window
and assigns the super-node a midpoint location, summed demand, and combined service
time. Levels repeat until $|V|\le P\,n_0$, where $n_0$ is the original number of
nodes, so the \emph{coarsening depth} $P\in(0,1]$ fixes the target fraction of nodes
that survive: smaller $P$ means more aggressive merging and a smaller QUBO.
\emph{Inflation} reverses the recorded merges to recover a solution on the original
graph (Fig.~\ref{fig:coarsening}). In the baseline, the four hyperparameters
$(\alpha,\beta,P,\text{radiusCoeff})$ are tuned separately for each family.

\begin{algorithm}[htbp]
\SetAlgoLined\DontPrintSemicolon
\KwIn{graph $G$, params $\alpha,\beta,P,\rho\text{-coeff}$}
\While{$|V|>P\,n_0$}{
  score every edge $e.\Dij \leftarrow \text{scorer}(G,e)$; sort ascending\;
  $\rho \leftarrow \Dij$ at index $\lfloor 0.1\,|E|\,\rho\text{-coeff}\rfloor$;
  $U\leftarrow\{\text{depot}\}$\;
  \ForEach{edge $(i,j)$, $\Dij\le\rho$, $i,j\notin U$}{
    \lIf{neither $e_i{+}s_i{+}\tau_{ij}\le l_j$ nor $e_j{+}s_j{+}\tau_{ij}\le l_i$}{skip}
    choose order by larger slack; window $[e',l']$ (skip if $l'<e'$); record; $U\!\mathrel{+}\!=\!\{i,j\}$\;
  }
  \lIf{no merges recorded}{break} build super-nodes (Eq.~\eqref{eq:supernode});
  reconnect; remove merged nodes\;
}
\caption{One level of spatio-temporal coarsening (scorer = heuristic $\Dij$ or GNN)}
\label{alg:coarsen}
\end{algorithm}

Merging $i$ and $j$ (in the time-feasible order $i\!\to\!j$) creates a super-node
$ij$ whose attributes are
\begin{equation}
\left(x_{ij},y_{ij}\right)=\left(\tfrac{x_i+x_j}{2},\,\tfrac{y_i+y_j}{2}\right),
\quad
d_{ij}=d_i+d_j,
\quad
s_{ij}=s_i+\tau_{ij}+s_j,
\quad
[e_{ij},l_{ij}]=[e',l'],
\label{eq:supernode}
\end{equation}
where $[e',l']$ is the time window implied by the selected order. Including the
internal leg $\tau_{ij}$ in the service time prevents the coarse graph from counting
that travel again and preserves distance consistency after inflation. Demands are
summed because both customers must be served by the same vehicle. Inflation restores
$i$ and $j$ in the recorded order.

\subsection{QUBO formulation}
\label{sec:method-qubo}
We use the position encoding of Borowski et al.\ \citep{borowski2020}, in which the
binary variable $x_{i,j,k}=1$ if and only if vehicle $i$ visits customer $j$ at
route step $k$. A route is an ordered sequence of occupied steps, and one assignment
of these variables describes the whole fleet's schedule. For an instance with $n_c$
customers the fleet size is $n_v=\max(2,\lfloor n_c/2\rfloor)$ and each vehicle is
allotted $k_{\max}$ route steps, giving $n_v\,n_c\,k_{\max}$ position variables.

The two solvers of \citet{borowski2020} differ only in $k_{\max}$. The Full QUBO
Solver (FQS) gives every vehicle $\lceil n_c/n_v\rceil+1$ steps, one more than an
even share of the customers, and solves the resulting QUBO in a single pass. The
Average Partitioning Solver (APS) gives $\lceil n_c/n_v\rceil+r$ steps for a radius
parameter $r$, trading problem size against the freedom to depart from an even
split. At the default $r{=}1$ the two prescriptions coincide, which is the
configuration used throughout this paper; we therefore write FQS/APS for the common
encoding and set $k_{\max}=\min(\lceil n_c/n_v\rceil+1,\,n_c)$.

Constraints enter as weighted penalty terms, each of which vanishes when the
constraint is satisfied:
\begin{itemize}[nosep]
  \item \textbf{visit-uniqueness}: each customer occupies exactly one
  (vehicle, step) slot, and no vehicle occupies two customers at the same step;
  \item \textbf{capacity}: the demand carried by a vehicle may not exceed $Q$,
  encoded as Eq.~\eqref{eq:capslack} below;
  \item \textbf{self-loop}: a vehicle may not remain at the same customer over two
  consecutive steps;
  \item \textbf{time-window}: service at customer $j$ must begin within
  $[e_j,l_j]$;
  \item \textbf{continuity}: a vehicle may not use step $k$ without also using step
  $k-1$, so a route is contiguous and cannot contain gaps;
  \item \textbf{distance}: the routing objective, a Clarke--Wright savings score
  \citep{clarke1964}. This is the quantity being minimised rather than a
  constraint, and it carries the smallest weight.
\end{itemize}
The weights decrease strictly in the order visit-uniqueness $>$ capacity, self-loop
$>$ time-window $>$ continuity $>$ distance, so that violating a constraint can
never be offset by a gain in a weaker one. Capacity and self-loop are distinct
constraints that share a weight tier. The time-window penalty has three parts: a
depot-reachability check, a pairwise term between consecutive steps, and a triangle
look-ahead spanning steps $k$ to $k+2$. Together these exclude orderings that cannot
be served in time, without enumerating complete routes.

Following the coarsening-QUBO implementation we build on, capacity is expressed as
an equality using binary log-slack. For vehicle $i$,
\begin{equation}
\sum_j d_j\sum_k x_{i,j,k}\;+\;\sum_{m=0}^{M-1} 2^{m}\,s_{i,m}\;=\;Q ,
\qquad M=\lfloor\log_2 Q\rfloor+1 ,
\label{eq:capslack}
\end{equation}
where the $s_{i,m}$ are auxiliary binary slack bits and $m$ indexes their powers of
two, so the slack term can represent any unused capacity between $0$ and $Q$.
Forming the penalty squares this expression, and with it the slack coefficients,
producing couplings as large as $\bigl(2^{M-1}\bigr)^{2}P_{\text{cap}}$ where
$P_{\text{cap}}$ is the capacity penalty weight. These couplings are central to
Section~\ref{sec:phase2}.

\subsection{Adaptive penalty calibration}
\label{sec:method-adaptive}
The previous paper \citep{ozyilmaz2025} uses fixed penalties regardless of the
instance scale. Choosing penalty weights is itself a studied problem: weights must
be large enough that no constraint violation is profitable, yet no larger, because
the finite coefficient precision of an annealer is spent on whatever range the
weights span \citep{ayodele2022penalty}. We therefore set penalties relative to the
instance's objective magnitude
\begin{equation}
B=w_{\mathrm{dist}}\cdot\max_{ij}c_{ij}\cdot n_c ,
\label{eq:Bdef}
\end{equation}
an upper bound on the total routing objective, where $c_{ij}$ is the Euclidean arc
cost between nodes $i$ and $j$, $n_c$ is the number of customers, and
$w_{\mathrm{dist}}$ is the weight given to the distance objective. Each constraint
weight is a fixed multiple of $B$: unique $=15B$, capacity $=\lambda_{\mathrm{cap}}B$
with $\lambda_{\mathrm{cap}}=8$, and time-window $=5B$, preserving the hierarchy of
Section~\ref{sec:method-qubo}. We additionally \emph{drop} any capacity constraint
that cannot bind (total demand $\le Q$) and, when it does bind, size the slack to
$\min(Q,\text{total demand})$ and raise $\lambda_{\mathrm{cap}}$ in proportion to the
demand pressure, the ratio of total demand to $Q$.

A binding constraint is precisely the case that survives the drop, and it carries
its log-slack with it. Its largest coupling scales as $(2^{m_{\max}})^2
P_{\text{cap}}$ with $m_{\max}=\lfloor\log_2 Q\rfloor$, so an unnormalised weight
places the capacity block orders of magnitude above every other term: the calibration
would inflate exactly the dynamic range it exists to control. We therefore normalise
the binding capacity weight by that squared slack coefficient,
\begin{equation}
P_{\text{cap}}=\frac{\lambda_{\mathrm{cap}}B}{\left(2^{\lfloor\log_2 Q\rfloor}\right)^{2}},
\label{eq:capnorm}
\end{equation}
which returns the capacity block to the scale of the remaining penalties while
preserving the hierarchy. The normalisation is inert when the constraint is dropped,
so it acts only on binding instances; the fraction of instances it touches grows with
$N$ (2 of 13 at $N{=}10$, 9 of 13 at $N{=}40$ on our test set). The scaling alone
changes little; the capacity handling is the lever (Section~\ref{sec:phase2}).

\subsection{GNN merge scorer}
\label{sec:method-gnn}
\begin{figure}[htbp]
\centering
\includegraphics[width=\textwidth]{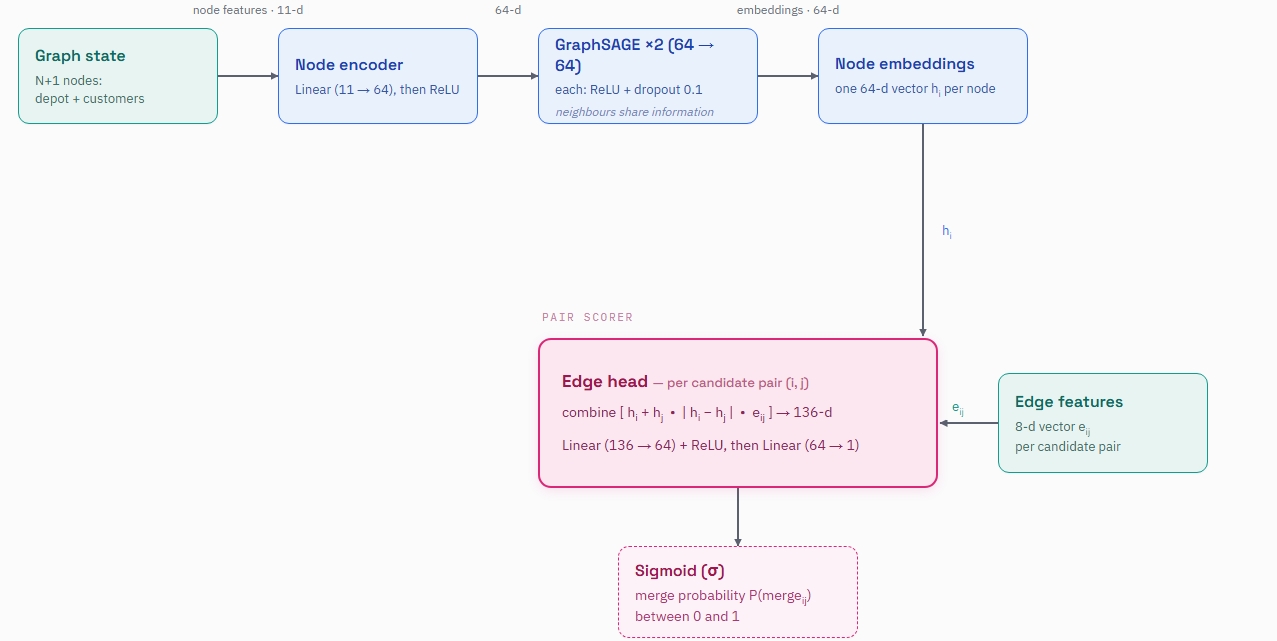}
\caption{The merge-scorer GNN. A message-passing trunk produces a 64-d embedding
$h_i$ per node; a symmetric edge head takes the node embeddings and the 8-d edge
features and outputs $P(\mathrm{merge}_{ij})=\sigma(\cdot)$, used in place of the
heuristic score $\Dij$.}
\label{fig:gnn}
\end{figure}

The learned scorer (Fig.~\ref{fig:gnn}) consists of a node encoder, two GraphSAGE
message-passing layers, and a symmetric edge head that assigns a merge probability
to each candidate pair. The 11-dimensional node features comprise coordinates,
demand, service time, $[e,l,t]$, window width, degree, a super-node indicator, and
distance to the depot. The eight-dimensional edge features comprise $\tau_{ij}$,
bidirectional temporal slack, window overlap, feasibility indicators, combined
demand, and $|t_i-t_j|$. The learned probability replaces $\Dij$ in
Algorithm~\ref{alg:coarsen}; all other steps, including the explicit feasibility
check, remain unchanged.

Training labels are obtained from an instance-specific \emph{oracle}: the
highest-reward outcome among five heuristic coarsenings from a fixed sweep over
$(\alpha,\beta,P,\text{radiusCoeff})$ and the current model's rollouts. Every
candidate is evaluated through the complete downstream pipeline. We record each
coarsening level, allowing the model to learn decisions on intermediate graphs that
already contain super-nodes, as it encounters during inference. Because the oracle
is selected by downstream performance rather than by family identity, its labels
combine successful merge decisions across families without exposing the family
label to the model.

\subsection{Reward-driven expert iteration}
\label{sec:method-ei}
Direct imitation of a heuristic-derived oracle cannot exceed the best behaviour in
the heuristic pool. We therefore augment the labels with the GNN's own
\emph{stochastic} coarsening rollouts. Independent Gaussian noise with standard
deviation $0.35$ is added to the edge scores before ranking, causing each rollout
to sample a different coarsening. Every candidate is evaluated end to end using

\[
r = 2\,[\text{feasible}] + [\text{pre-feasible}]
    - 0.5\min(\text{gap},2) - 0.02\min(v,50).
\]

The four quantities are measured on the solution a candidate coarsening produces
after the full pipeline has run. $[\text{feasible}]$ is $1$ when the final solution,
after the classical repair of Section~\ref{sec:method-decode}, satisfies every
constraint, and $0$ otherwise. $[\text{pre-feasible}]$ is the same indicator applied
to the decoded solution \emph{before} repair, so it measures the raw quality of the
sampled assignment. $\text{gap}$ is the relative cost excess over the OR-Tools
reference, $(c-c_{\mathrm{ref}})/c_{\mathrm{ref}}$, and $v$ is the number of
constraint violations in the same pre-repair solution.

The highest-reward candidate for each instance is designated the \emph{oracle}, the
worked example the model is trained to imitate on the next round
(Algorithm~\ref{alg:ei}). The
coefficients specify a priority order rather than a tuned optimum: post-repair
feasibility precedes pre-repair feasibility, which precedes cost and then raw
violations. The caps prevent one poor rollout from dominating the comparison. The
weights were selected manually and are used only to rank candidates within the same
instance. When a model rollout outperforms all heuristic settings, its decisions
enter the next training set, allowing the policy to improve beyond imitation of the
heuristic pool.

\begin{algorithm}[htbp]
\SetAlgoLined\DontPrintSemicolon
\KwIn{current model $g$, training instances}
\ForEach{instance}{
  pool $\leftarrow$ heuristic-setting coarsenings $\cup$ stochastic $g$-rollouts\;
  \lForEach{candidate}{run pipeline; reward $\leftarrow$ feasibility, then gap to OR-Tools, then violations}
  oracle $\leftarrow$ candidate with maximum reward; add its per-level merges to the labels\;
}
retrain $g$\;
\caption{Expert-iteration labelling (one round)}
\label{alg:ei}
\end{algorithm}

\subsection{Decoding and repair}
\label{sec:method-decode}
The lowest-energy sample is decoded to routes and passed through a classical repair
(duplicate removal, greedy insertion of missing customers, per-route time-window
repair, inter-route moves). This repair is strong and dominates post-repair
feasibility; we therefore also report \emph{pre-repair} metrics, which isolate the
raw solver/coarsening quality.

\section{Experimental setup}
\begin{figure}[htbp]
\centering
\includegraphics[width=\textwidth]{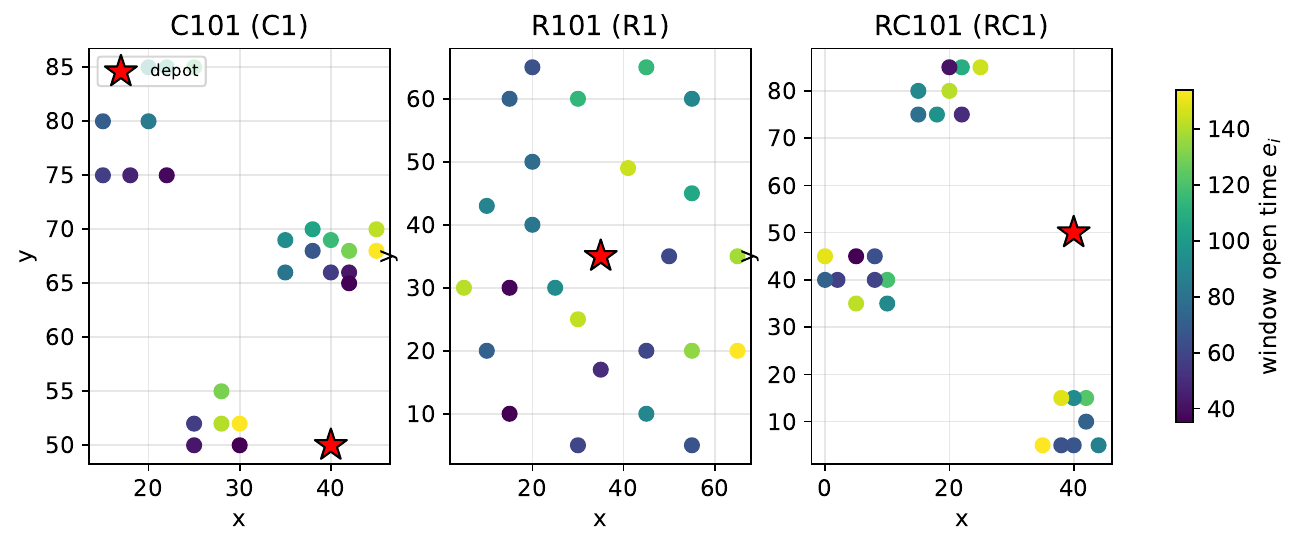}
\caption{Representative Solomon instances (first 25 customers); colour encodes the
window-open time $e_i$, and the red star is the depot. R (random) is the hard
family throughout.}
\label{fig:instances}
\end{figure}

\paragraph{Data.} The Solomon CVRPTW benchmark \citep{solomon1987} contains 56
instances in six families. The C, R, and RC designations denote clustered, random, and mixed customer
locations, respectively; each is paired with either a narrow or a wide scheduling
horizon. Series~1 has narrower time windows and lower vehicle capacity, so each
route typically serves fewer customers. Series~2 has wider windows and greater
capacity. Capacity constraints are therefore more likely to bind in the
tight-horizon families, which motivates the normalisation in
Eq.~\eqref{eq:capnorm} (Fig.~\ref{fig:instances}). Following prior work, we retain
the first $N$ customers for $N=5$ through the complete $N=100$ instance, thereby
preserving the family structure. GNN experiments use a deterministic,
family-stratified split of 43 training and 13 held-out test instances. The test set
is not used for training or model selection.

\paragraph{Reference solver.} OR-Tools runs a guided-local-search metaheuristic
under a fixed time limit and does not certify optimality, so we treat its cost as a
strong classical reference rather than a proven optimum. Accordingly, we report the
\emph{gap to the OR-Tools reference}, $(c-c_{\mathrm{ref}})/c_{\mathrm{ref}}$ with
$c_{\mathrm{ref}}$ the OR-Tools cost, over runs that are feasible, and avoid the
term ``optimality gap''. The per-family
heuristic uses the previous paper's published per-family hyperparameters, fixed in
prior work and never tuned on our test instances; the GNN uses a single fixed
configuration, so the comparison does not favour the GNN.

\paragraph{Solver and statistics.} All compared conditions receive the same solver
budget, including the backend, number of reads, sweeps, and seed. Each experiment
uses at least three seeds; the penalty and coarsening studies use five to ten. We
report bootstrap 95\% confidence intervals. Runs with different seeds on the same
benchmark instance are not independent problem observations. The primary Wilcoxon
signed-rank tests therefore take the median across seeds to obtain one value per
instance before forming pairs; the pooled per-seed test is included only as a
secondary reference. Cost-gap comparisons include pairs for which both methods are
feasible, with feasibility reported separately. Experiments run on a single
workstation, and the GPU is used only to train the GNN.

\section{Results}

\subsection{Baseline reproduction}
\begin{figure}[htbp]
\centering
\includegraphics[width=0.49\textwidth]{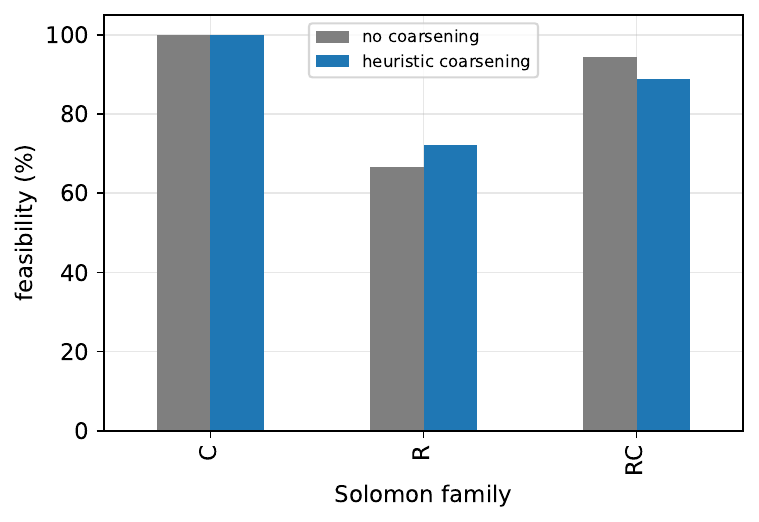}\hfill
\includegraphics[width=0.49\textwidth]{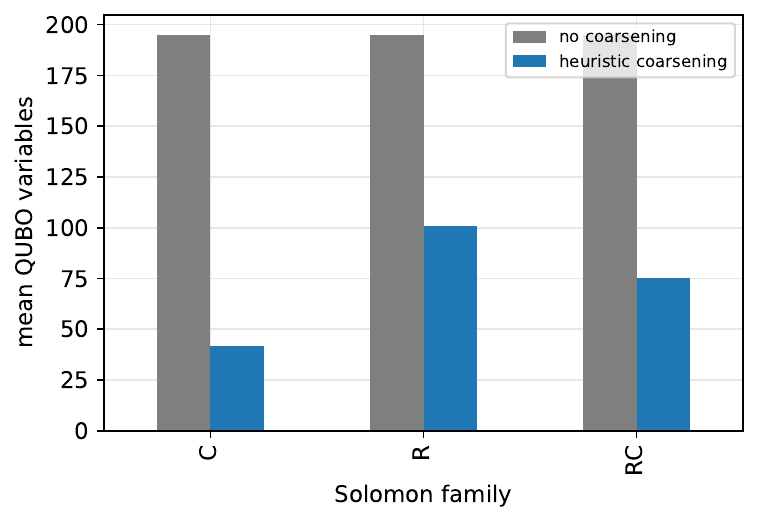}
\caption{Baseline ($N{=}10$, static penalties). Left: family-dependent feasibility
(R-type is the weak spot). Right: coarsening reduces QUBO size.}
\label{fig:phase1}
\end{figure}

At $N{=}10$, we reproduce the qualitative behaviour reported previously
(Fig.~\ref{fig:phase1}). All decoded solutions are valid. Feasibility is $100\%$
for C-type instances, falls to $66.7\%$ without coarsening and $72.2\%$ with
heuristic coarsening for R-type instances, and lies between these values for RC.
Coarsening reduces the QUBO size by approximately $P^2$, accelerates sampling by a
factor of about $3.2$, and lowers route distance by $16$--$25\%$ relative to the
uncoarsened formulation among feasible runs.

\subsection{Adaptive penalty calibration}
\label{sec:phase2}
\begin{figure}[htbp]
\centering
\includegraphics[width=0.32\textwidth]{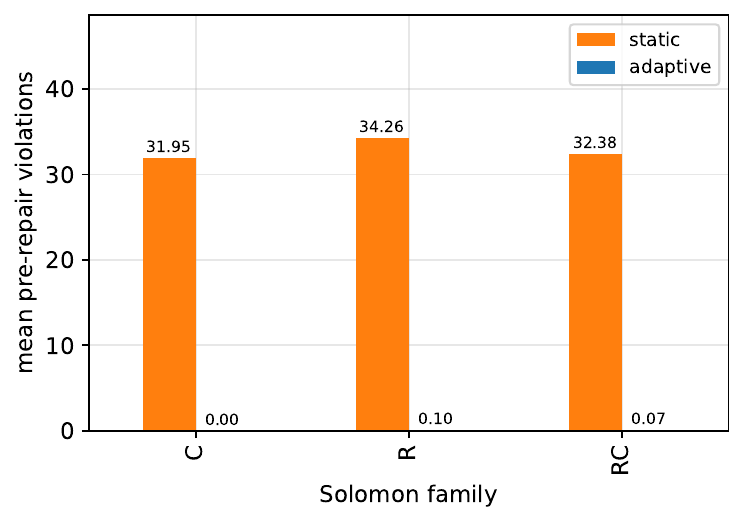}\hfill
\includegraphics[width=0.32\textwidth]{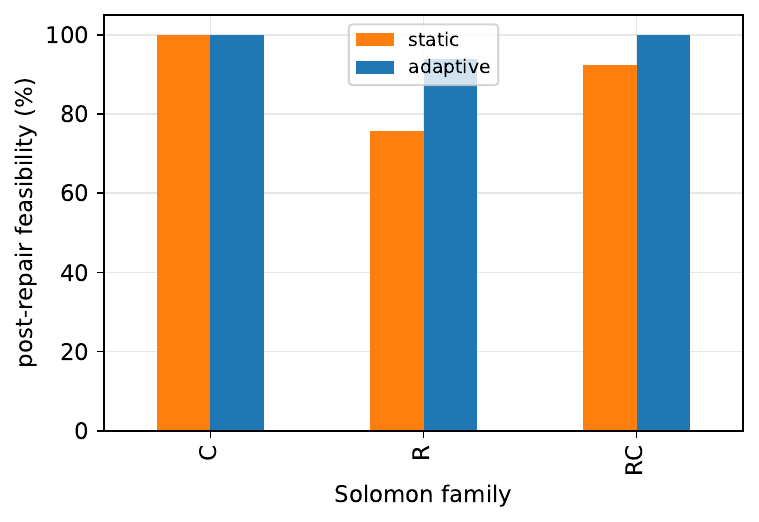}\hfill
\includegraphics[width=0.32\textwidth]{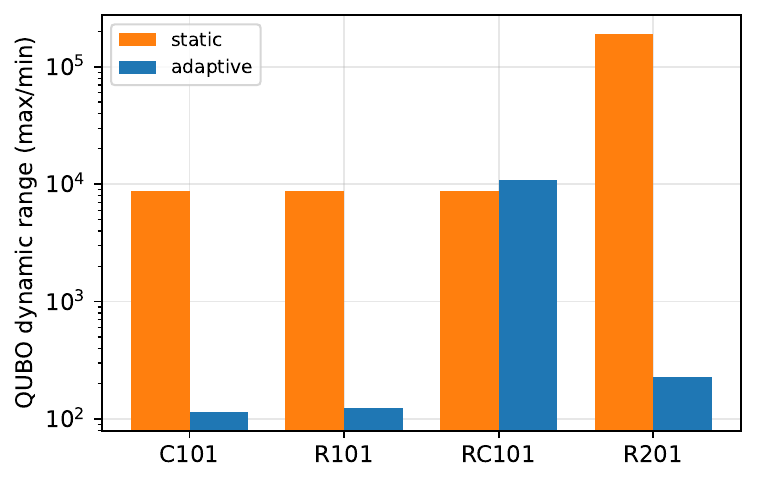}
\caption{Adaptive penalties. (a) Pre-repair violations by family. (b) Post-repair
feasibility. (c) The internal dynamic range collapses by orders of magnitude for
non-binding families (log scale); RC1 binds, so its capacity constraint is retained.}
\label{fig:phase2}
\end{figure}

\textbf{Conditioning mechanism.} Uniform scaling has almost no effect under
simulated annealing because the sampler derives its temperature schedule from the
coupling magnitudes. Rescaling all coefficients therefore leaves the effective
sampling distribution nearly unchanged. The relevant quantity is the QUBO's
internal coefficient range, which is dominated by the capacity log-slack couplings
of Eq.~\eqref{eq:capslack}. Squaring the slack term makes these of order
$(2^{m})^{2}P_{\text{cap}}$, where $m$ indexes the slack bits and so runs up to
$\lfloor\log_2 Q\rfloor$; for $Q{=}1000$ the largest such coupling is
approximately $10^{12}$. At
$N{=}10$, the capacity constraint does not bind in five of the six families, so
these large couplings encode a redundant constraint. Removing it reduces the
coefficient range from $8.7\!\times\!10^3$ to $115$ for C101 and from
$1.9\!\times\!10^5$ to $228$ for R201 (Fig.~\ref{fig:phase2}c).

\textbf{Effect.} Under an equal solver budget (56 instances and five seeds;
Table~\ref{tab:phase2}, Fig.~\ref{fig:phase2}), adaptive calibration reduces the
mean number of pre-repair violations from
$33.0$ to $0.06$, lifts pre-repair feasibility from $0.7\%$ to $94.8\%$, and
improves post-repair feasibility. A paired Wilcoxon test on pre-repair violations,
aggregated to one value per instance, gives $p\!=\!3.7\times10^{-11}$ ($n=56$; the
seed-pooled value is $4.4\times10^{-93}$, $n=560$). With a binding-aware capacity
weight and the normalisation of Eq.~\eqref{eq:capnorm}, adaptive is at least as good
as static on every family group, including the binding RC1 group ($100\%$ vs.\ $85\%$ post-repair).
The same principle applies to other slack-encoded constraints: remove the
constraint when it cannot bind, and normalise its coefficient block when it can.

\begin{table}[htbp]
\centering\small
\caption{Adaptive penalty calibration ($N{=}10$, equal budget, 56 instances
$\times$ 5 seeds $\times$ 2 coarsening conditions). Paired Wilcoxon on pre-repair
violations, aggregated to one value per instance: $p=3.7\times10^{-11}$,
$n=56$ (seed-pooled: $p=4.4\times10^{-93}$, $n=560$).}
\label{tab:phase2}
\begin{tabular}{lcccc}
\toprule
penalties & pre-repair viol. & pre-repair feas. & post-repair feas. & QUBO vars \\ \midrule
static & 33.0 & 0.7\% & 87.9\% & 131 \\
\textbf{adaptive} & \textbf{0.06} & \textbf{94.8\%} & \textbf{97.5\%} & \textbf{101} \\
\bottomrule
\end{tabular}
\end{table}

\begin{figure}[htbp]
\centering
\includegraphics[width=0.9\textwidth]{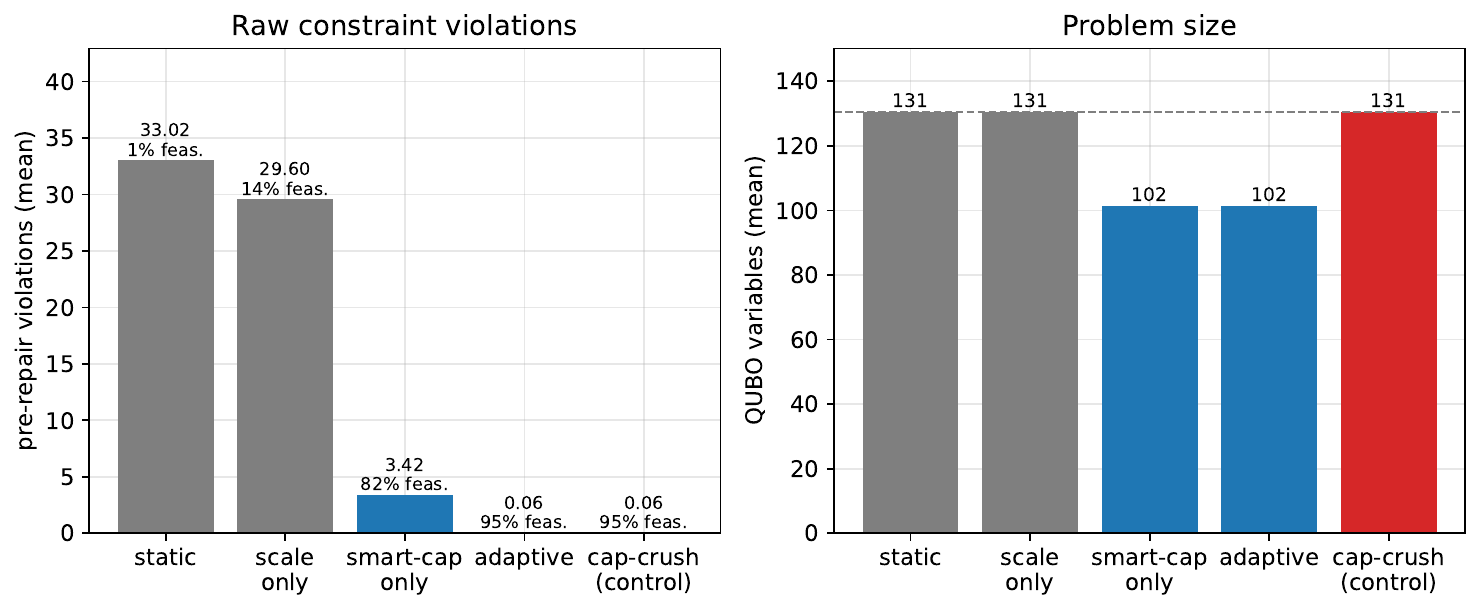}
\caption{Lever ablation. Left: pre-repair violations (with feasibility annotated);
right: QUBO variable count. Objective-scaling alone is nearly inert. Smart-cap
collapses violations but also removes slack variables ($131\!\to\!101$), and leaves a
residual on binding instances that normalising the binding capacity weight removes.
The \emph{cap-crush} control keeps the static variable count ($131$) yet, by
collapsing only the dynamic range, matches full adaptive, isolating conditioning
from problem size.}
\label{fig:ablation}
\end{figure}

\textbf{Ablation.} Decoupling the levers (Fig.~\ref{fig:ablation}) separates the
effect into three steps. Objective-scaling the weights alone (\emph{scale-only})
moves pre-repair violations only from $33.0$ to $29.6$ (instance-level
$p\!=\!0.03$): a marginal shift, two orders of magnitude short of the other levers,
confirming that uniform magnitude is nearly inert. Dropping non-binding capacity
alone (\emph{smart-cap only}) does most of the work ($33.0\!\to\!3.4$,
$p\!=\!8.1\times10^{-10}$). Normalising the weight of the constraints that
\emph{do} bind (Eq.~\eqref{eq:capnorm}) closes the remaining gap, taking full
adaptive to $0.06$ violations and $94.8\%$ pre-repair feasibility: the residual
$3.4$ of smart-cap is concentrated entirely on binding instances, where the slack
survives and its $(2^{m})^2$ couplings still dominate.

Dropping the constraint, however, removes its slack variables as well
($131\!\to\!101$), so smart-cap conflates dynamic-range conditioning with a smaller
search space. We therefore add a control, \emph{cap-crush}, that keeps every
capacity variable but crushes the capacity penalty so its $(2^m)^2$ couplings no
longer dominate. At the static variable count ($131$), cap-crush matches full
adaptive (pre-repair violations $0.06$, feasibility $94.6\%$; instance-level
$p\!=\!3.7\times10^{-11}$, $n=56$) while using $30$ more variables. Because it
changes only the dynamic range and not the variable count, this isolates the
mechanism: the lever is the QUBO's internal dynamic range (conditioning), not the
smaller search space. The capacity-aware handling, not uniform scaling, drives it.

\subsection{GNN-guided, tuning-free coarsening}
\label{sec:phase3}
\begin{figure}[htbp]
\centering
\includegraphics[width=0.85\textwidth]{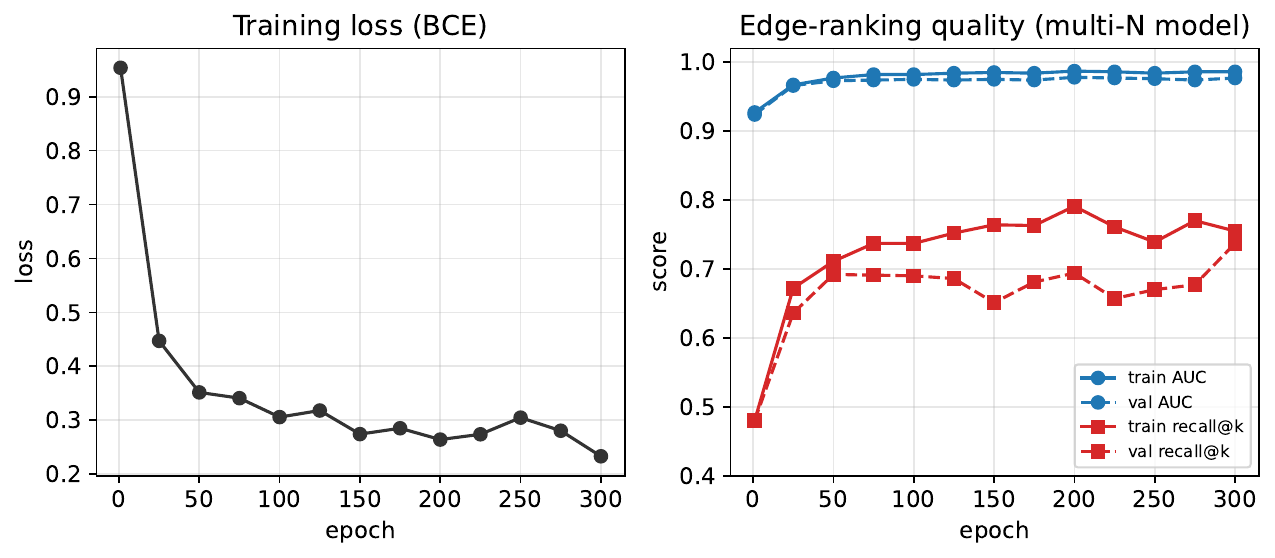}
\caption{GNN training (multi-size model). Left: BCE loss. Right: edge-ranking AUC
and recall@$k$ on train and on held-out-instance validation.}
\label{fig:train}
\end{figure}

\begin{figure}[htbp]
\centering
\includegraphics[width=0.5\textwidth]{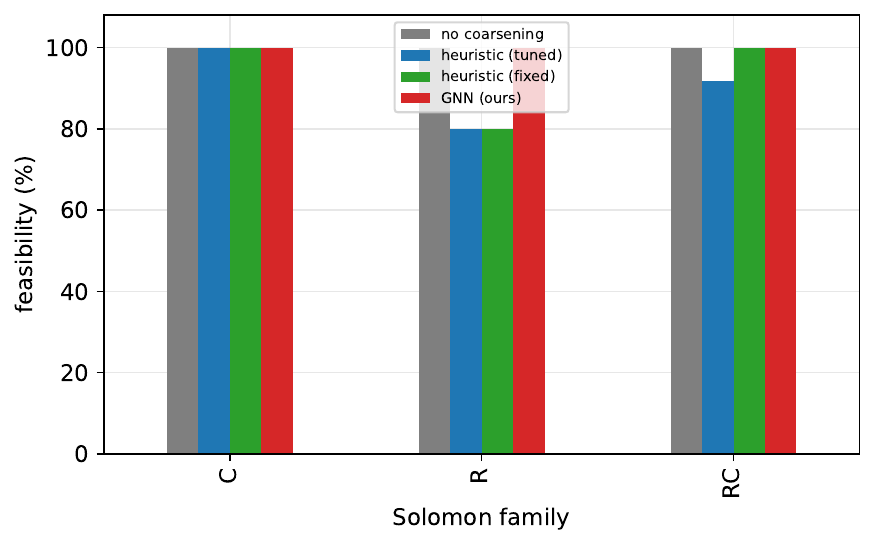}
\caption{$N{=}10$ feasibility by family. The GNN, with no per-family tuning,
matches or beats the per-family-tuned heuristic, notably on R-type.}
\label{fig:phase3}
\end{figure}

The GNN is trained on $N\in\{10,15,20\}$ (Fig.~\ref{fig:train}) and attains an
edge-ranking AUC of approximately $0.98$ on held-out instances. Here, AUC is the
probability that the model ranks an oracle merge above a non-merge candidate. With
one fixed configuration across all families, the model reaches $100\%$ feasibility
at $N{=}10$, including \textbf{$100\%$ on R-type instances compared with $80\%$
for the tuned heuristic}. It also produces the fewest raw violations and the
greatest compression (Fig.~\ref{fig:phase3}). The following section evaluates the
more informative question of whether this behaviour persists across problem sizes.

\subsection{Learned coarsening is feasibility-preserving}
\label{sec:phase4}
The learned scorer is intended to preserve feasibility without family-specific
tuning. To isolate the effect of the score, we compare it with
\emph{heuristic-fixed}, which uses the same coarsening settings as the GNN
($P{=}0.5$, radiusCoeff${=}0.5$). The merge score, either the fixed $\Dij$ or the
learned one, is therefore the only difference. Across the held-out test instances at
$N\in\{10,20,40,50,60,80,100\}$, the GNN's coarsened solutions are feasible more
often than the per-family-tuned heuristic's: $83.3\%$ vs.\ $68.9\%$ overall, better
or tied on $85$ of $90$ instance-size pairs (strictly better on $26$, worse on $5$;
paired Wilcoxon over those pairs $p\!=\!3.4\times10^{-4}$), using a single fixed
configuration for all families.

\begin{figure}[htbp]
\centering
\includegraphics[width=0.62\textwidth]{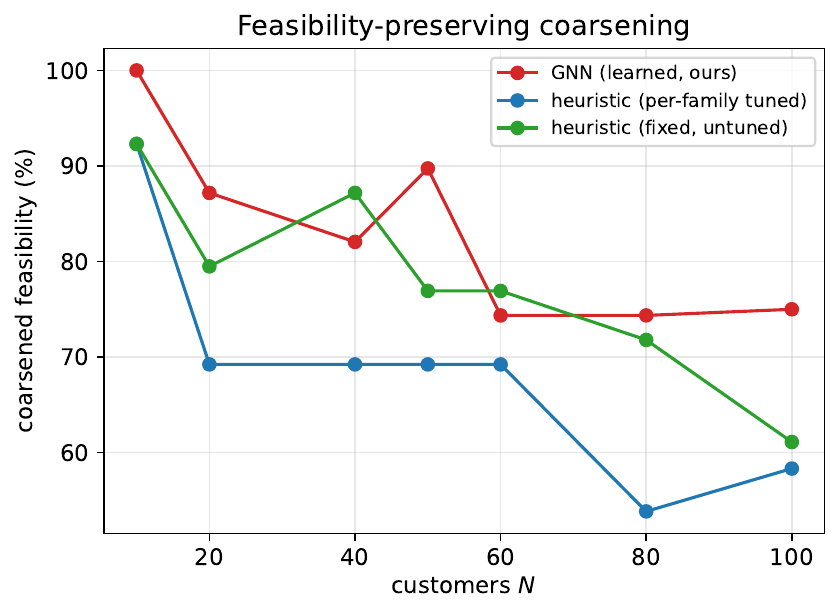}
\caption{Coarsened feasibility vs.\ $N$. The learned coarsener matches or beats the
per-family-tuned heuristic at every size, tuning-free; the gap widens with $N$ and is
significant at $N{=}80,100$ ($p{=}0.002$, better or tied on $25/25$). The heuristic's
per-family parameters were tuned at $N{=}10$ and do not transfer, so at large $N$ they
over-merge and it falls below even an untuned default, while the GNN, conditioning on
each instance, degrades gracefully.}
\label{fig:feasscale}
\end{figure}

The $90$ pairs are not mutually independent because each base instance appears at
seven sizes. Aggregating over size to obtain one observation for each of the $13$
independent base instances preserves the direction of the effect but gives
$p\!=\!0.055$. We therefore present the statistical comparison primarily at each
size, where the 13 instances are independent. At small $N$, both methods are near
the feasibility ceiling and their difference is not significant
(Fig.~\ref{fig:feasscale}). At $N{=}80,100$, the GNN achieves $74.7\%$ feasibility
compared with $56.0\%$ for the heuristic and is better or tied on all $25$ available
pairs ($p\!=\!0.002$).

The difference consequently increases with $N$. The family-specific heuristic
parameters were tuned at $N{=}10$ and transfer poorly to larger instances. At
$N{=}80,100$, the tuned heuristic falls to $53.8$--$58.3\%$ feasibility, below the
untuned fixed configuration, whereas the GNN remains at $74.4$--$75.0\%$. The
learned policy thus removes family-specific tuning and degrades more gradually
outside the regime in which the heuristic was calibrated.

\textbf{Classical repair bounds the final cost.} Expert iteration reduces the gap
to OR-Tools among the coarsening variants (Table~\ref{tab:phase4}); the model's own
rollouts supply the oracle in $46\%$ of the training cases. A purely classical
baseline nevertheless outperforms every QUBO variant. Applying the same repair to
an empty solution (\emph{repair-only}: greedy insertion without SA, QUBO, or
coarsening) produces $100\%$ feasibility with a $16$--$41\%$ cost gap. Adding the
shared local search (2-opt and relocate) reduces this range to $5$--$19\%$. The
same local search brings the GNN pipeline to the level of, but not below, this
baseline: the mean gaps are $16.4\%$ and $13.2\%$, respectively; the GNN is lower
on $18$ of $38$ instance--size cells, with paired $p\!=\!0.72$. Our claims
therefore concern QUBO formulation and preprocessing rather than end-to-end cost.
The role of coarsening is to improve tractability and tuning-free feasibility.

\begin{table}[htbp]
\centering\small
\caption{Gap to the OR-Tools reference (\%, mean over feasible runs; 13 held-out
instances). Every quantum-pipeline variant (top) sits above the classical repair
baselines (bottom); a classical greedy repair with local search is cheapest.}
\label{tab:phase4}
\begin{tabular}{lccc}
\toprule
method & $N{=}10$ & $N{=}15$ & $N{=}20$ \\ \midrule
no coarsening (QUBO) & 149.7 & 184.7 & 232.5 \\
heuristic coarsening (QUBO) & 43.4 & 65.4 & 92.3 \\
GNN coarsening (QUBO, ours) & 35.8 & 52.4 & 64.9 \\ \midrule
classical repair-only & 16.3 & 33.4 & 40.8 \\
classical repair-only $+$ local search & \textbf{4.7} & \textbf{16.3} & \textbf{18.5} \\
\bottomrule
\end{tabular}
\end{table}

\paragraph{Convergence.} A second expert-iteration round does not improve the
policy (oracle-win rate $46\%\!\to\!39\%$), so the training converges in a single
round.

\subsection{Tractability and scaling to $N{=}100$}
\label{sec:scaling}
\begin{figure}[htbp]
\centering
\includegraphics[width=0.62\textwidth]{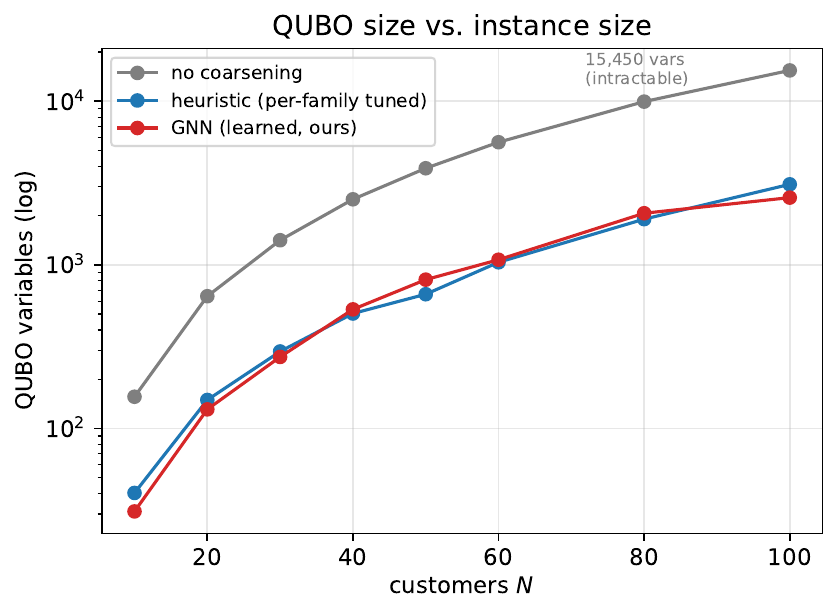}
\caption{QUBO variable count vs.\ $N$ (log scale). The uncoarsened encoding explodes
to $\sim\!15{,}000$ variables at $N{=}100$; coarsening keeps it $\sim\!5$--$6\times$
smaller and solvable throughout. Embeddability on current hardware is bounded far
lower, at $N\approx20$ (Section~\ref{sec:embed}).}
\label{fig:scalingvars}
\end{figure}

The principal value of coarsening is tractability. The uncoarsened FQS/APS QUBO grows
quickly with $N$, from $156$ variables at $N{=}10$ to $2{,}517$ at $N{=}40$ and
$15{,}450$ at $N{=}100$ (Fig.~\ref{fig:scalingvars}). Adaptive conditioning is
strong enough that simulated annealing still solves the uncoarsened problem at
$N{=}40$, but it does so on five times as many variables and at roughly three times
the cost per solve ($2.2$\,s vs.\ $0.8$\,s), and the resulting QUBO is far beyond
what a current annealer can embed (Section~\ref{sec:embed}). Coarsening holds the QUBO
$\approx\!5$--$6\times$ smaller ($31$, $534$, $2{,}577$ variables at $N{=}10,40,100$),
keeping the reduced problem solvable through $N{=}100$. This gain is a constant
factor: the coarsened QUBO still grows with $N$. The remaining growth motivates the
learned-depth experiment in Section~\ref{sec:learnedstop} and limits hardware
embeddability at a much smaller scale than classical solvability
(Section~\ref{sec:embed}). Because classical repair also makes the uncoarsened
pipeline feasible after repair, coarsening is needed for QUBO size, solvability,
and embeddability rather than for post-repair feasibility.

\subsection{Learned stopping}
\label{sec:learnedstop}
\begin{figure}[htbp]
\centering
\includegraphics[width=0.9\textwidth]{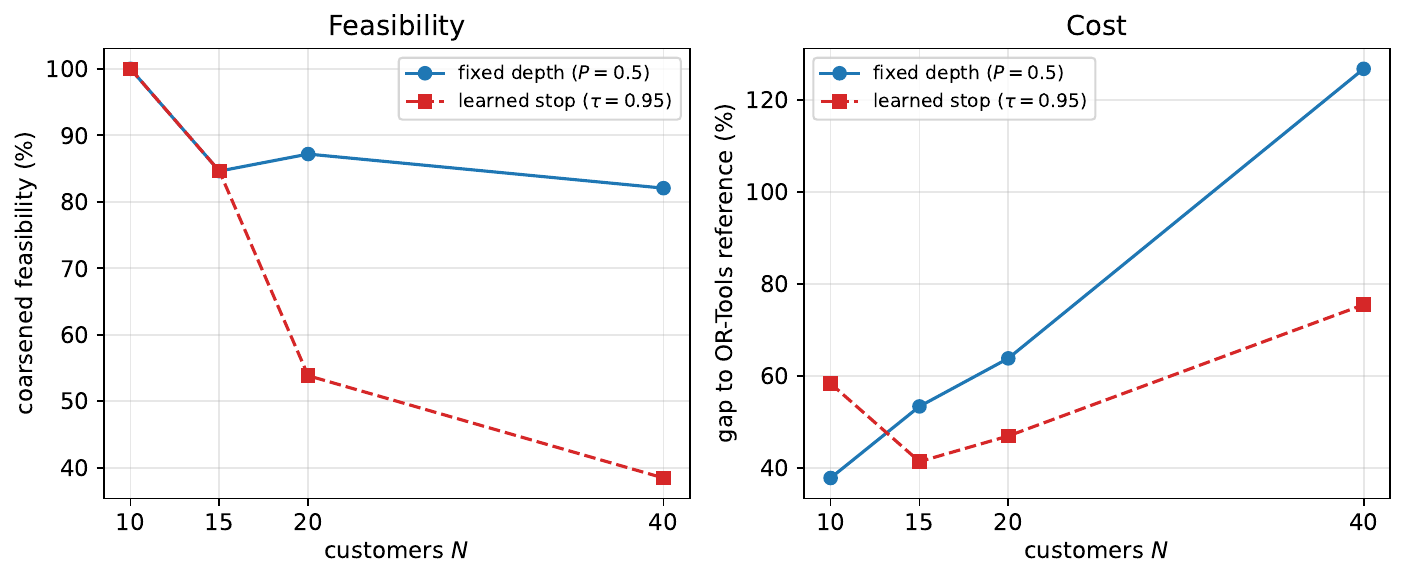}
\caption{Learned stopping trades feasibility for cost rather than dominating a fixed
coarsening depth.}
\label{fig:learnedstop}
\end{figure}

The fixed coarsening depth contributes to the decline in feasibility at larger $N$
(Fig.~\ref{fig:feasscale}). We test a stopping rule that terminates coarsening when
the highest predicted merge probability falls below a threshold $\tau$. The rule
does not dominate a fixed depth; instead, it exchanges feasibility for cost
(Fig.~\ref{fig:learnedstop}). At $N{=}40$, for example, the gap decreases from
$127\%$ to $76\%$, while feasibility falls from $82\%$ to $39\%$. Predicted merge
probabilities are nearly binary, preventing one global threshold from controlling
the depth finely. A context-dependent stopping head may provide a more flexible
criterion.

\subsection{Robustness: architecture and reward sensitivity}
\label{sec:robust}
Two design choices could in principle account for the reported gains rather than the
mechanisms we identify: the GraphSAGE trunk of the merge scorer, and the
coefficients of the expert-iteration reward. We examine both, holding everything
else fixed.

\textbf{Message-passing architecture.} We retrain the scorer with a GAT and a GCN
trunk in place of GraphSAGE under identical features, labels, optimiser, epochs and
validation protocol, then evaluate each by running the full pipeline on the held-out
instances. A single training run cannot separate an architecture effect from
initialisation noise, so each variant is trained three times and we report the
spread across those runs (Table~\ref{tab:arch}). No trunk is consistently superior:
GraphSAGE leads by more than twice the seed-to-seed spread only at $N{=}20$, GAT
matches it at $N{=}10$, and GCN is marginally ahead at $N{=}40$ using $31\%$ fewer
parameters. Edge-ranking quality separates GAT from the other two but does not order
GraphSAGE and GCN, which reinforces the observation of Section~\ref{sec:phase4} that
a merge scorer is best assessed by its effect on the pipeline rather than by its
ranking accuracy. We therefore do not attribute the results to the choice of trunk.

\begin{table}[htbp]
\centering\small
\caption{Message-passing trunk ablation. Coarsened feasibility (\%) on the held-out
test instances, mean over three training seeds with the spread across those runs;
the variants are otherwise identical. No trunk is consistently better.}
\label{tab:arch}
\begin{tabular}{lccc}
\toprule
 & GraphSAGE (ours) & GAT & GCN \\ \midrule
$N{=}10$ & $87.2\pm3.6$ & $89.7\pm3.6$ & $89.7\pm7.3$ \\
$N{=}20$ & $\mathbf{88.0\pm2.4}$ & $77.8\pm4.8$ & $82.9\pm6.0$ \\
$N{=}40$ & $84.6\pm8.4$ & $74.4\pm3.6$ & $85.5\pm2.4$ \\ \midrule
validation AUC & $0.966\pm0.001$ & $0.948\pm0.010$ & $0.965\pm0.001$ \\
parameters & $26{,}113$ & $43{,}521$ & $17{,}921$ \\
\bottomrule
\end{tabular}
\end{table}

\textbf{Reward coefficients.} The reward of Section~\ref{sec:method-ei} ranks
candidates within an instance, so its influence is confined to which candidate it
selects. Caching each candidate's measured outcome once allows alternative
weightings to be evaluated exactly by re-scoring, without repeating the pipeline.
The selection proves almost invariant (Table~\ref{tab:reward}): raising the gap
weight fourfold changes the selected oracle on $3$ of $129$ instance-size pairs, and
raising the violation weight tenfold changes none. Only the removal of both trailing
terms alters it materially, changing $74$ of $129$ selections and reducing the rate
at which the model's own rollouts win the oracle slot from $35\%$ to $2\%$, which
reduces the procedure to imitation of the heuristic pool.

As this is the only setting whose training data differs substantially, we evaluate
it end-to-end with three training seeds per configuration. The resulting models are
indistinguishable: $89.7\%$ against $89.7\%$ coarsened feasibility at $N{=}10$, and
$82.1\%$ against $82.1\%$ at $N{=}20$, both differences well within the seed spread
($\pm3.6$ and $\pm2.1$ respectively). The coefficients therefore encode a priority
ordering to which the procedure is insensitive, and optimising them would not be
expected to improve the policy. This was measured at $N{=}10$ and $N{=}20$, where
both configurations lie close to the feasibility ceiling; whether a difference
emerges at larger $N$ was not tested.

\begin{table}[htbp]
\centering\small
\caption{Reward sensitivity. How often an alternative weighting selects a different
oracle, out of $129$ instance-size pairs, computed exactly by re-scoring cached
candidates; and the fraction of oracle slots won by the model's own rollouts. Only
removing both trailing terms changes the selection materially, and even then the
trained model is unchanged.}
\label{tab:reward}
\begin{tabular}{lcc}
\toprule
weighting $(w_{\text{feas}},w_{\text{pre}},w_{\text{gap}},w_{\text{viol}})$
 & oracle changes & rollouts win oracle \\ \midrule
$(2,\,1,\,0.5,\,0.02)$ \ published & --- & $35\%$ \\
$(2,\,1,\,0.5,\,0.20)$ \ violation heavy & $0/129$ & $35\%$ \\
$(1,\,2,\,0.5,\,0.02)$ \ pre-repair first & $2/129$ & $34\%$ \\
$(2,\,1,\,2.0,\,0.02)$ \ gap heavy & $3/129$ & $36\%$ \\
$(1,\,1,\,0.5,\,0.02)$ \ flat feasibility & $4/129$ & $36\%$ \\
$(1,\,0.5,\,2.0,\,0.02)$ \ cost first & $8/129$ & $37\%$ \\
$(2,\,1,\,0,\,0)$ \ no trailing terms & $74/129$ & $2\%$ \\
\bottomrule
\end{tabular}
\end{table}

\FloatBarrier
\section{Evaluation on quantum hardware}
\label{sec:qpu}
The preceding experiments use a classical annealing simulator. This separates the
algorithmic effect from device noise, but does not establish whether the
conditioning mechanism persists on physical hardware, where minor embedding and
finite coupler precision affect the sampled problem. We first use an offline study
to identify the embeddable problem sizes and then repeat the conditioning
comparison on a D-Wave Advantage2 processor.

\subsection{Embeddability}
\label{sec:embed}
\begin{figure}[htbp]
\centering
\includegraphics[width=\textwidth]{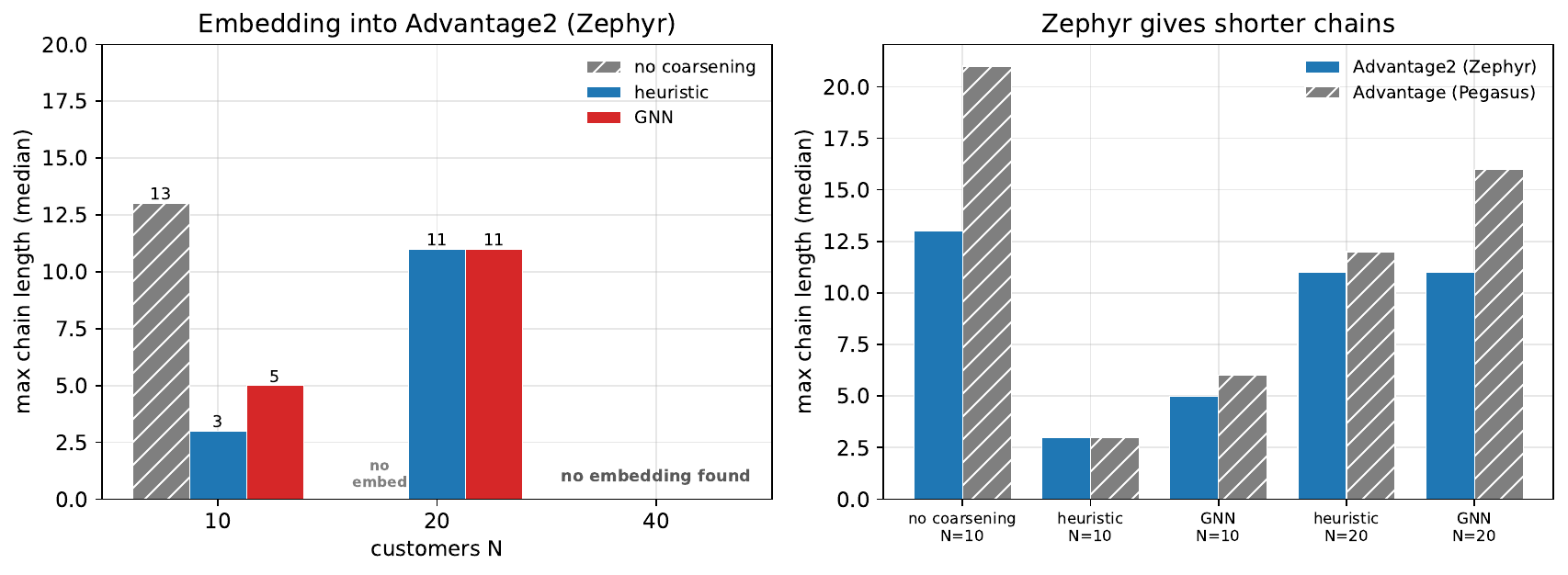}
\caption{Offline minor-embedding, no QPU time. Left: median maximum chain length on
Advantage2; the uncoarsened encoding fails to embed from $N{=}20$, and no variant
embeds at $N{=}40$. Right: Zephyr yields shorter chains than Pegasus wherever both
embed. Targets are full-yield ideal graphs, so these are optimistic bounds.}
\label{fig:embed}
\end{figure}

Before annealing, the logical QUBO must be minor-embedded in the processor's fixed,
sparse connectivity graph. A logical variable is then represented by a \emph{chain}
of physical qubits coupled ferromagnetically. Because longer chains are more prone
to breaks, the maximum chain length is often more restrictive in practice than the
total number of physical qubits. We embed the adaptive QUBO with
\code{minorminer} in ideal Pegasus (Advantage, $5{,}640$ qubits) and Zephyr
(Advantage2, $7{,}440$ qubits) graphs. Each cell contains three instances and three
seeds, and the study consumes no QPU time (Fig.~\ref{fig:embed}).

Coarsening determines whether these problems can be embedded. At $N{=}10$, the
uncoarsened QUBO embeds with a median maximum chain length of $13$ on Zephyr and
$21$ on Pegasus; the corresponding values are $3$ for heuristic coarsening and $5$
for GNN coarsening. At $N{=}20$, no embedding of the uncoarsened formulation
($653$ variables) is found on either topology within the search budget, whereas
both coarsened variants embed with maximum chains of length $11$. No variant embeds
at $N{=}40$. These results place the practical operating regime of the pipeline at
approximately $N{=}10$--$20$. Zephyr also produces consistently shorter chains
than Pegasus wherever both admit an embedding, motivating our use of Advantage2.
This embeddability benefit is attributable to coarsening in general, rather than to
the learned scorer specifically. The heuristic merges more aggressively and
therefore produces slightly smaller problems; the GNN's measured advantage concerns
feasibility and cost (Section~\ref{sec:phase4}), not embeddability.

\subsection{Conditioning on the annealer}
We evaluated the three penalty configurations from Section~\ref{sec:phase2} on
\code{Advantage2\_system1} (Zephyr topology, working graph \code{01138bbada}) over
the $13$ held-out test instances at $N{=}10$. We used GNN coarsening, $1{,}000$
reads, three seeds per cell, a separate minor embedding for each arm, and no gauge
averaging. Chain-break fractions remained small (median $0.011$, maximum $0.104$),
indicating that broken chains were not the principal source of differences among
the formulations. The complete study required less than $25$ seconds of QPU access
time.

\begin{figure}[htbp]
\centering
\includegraphics[width=\textwidth]{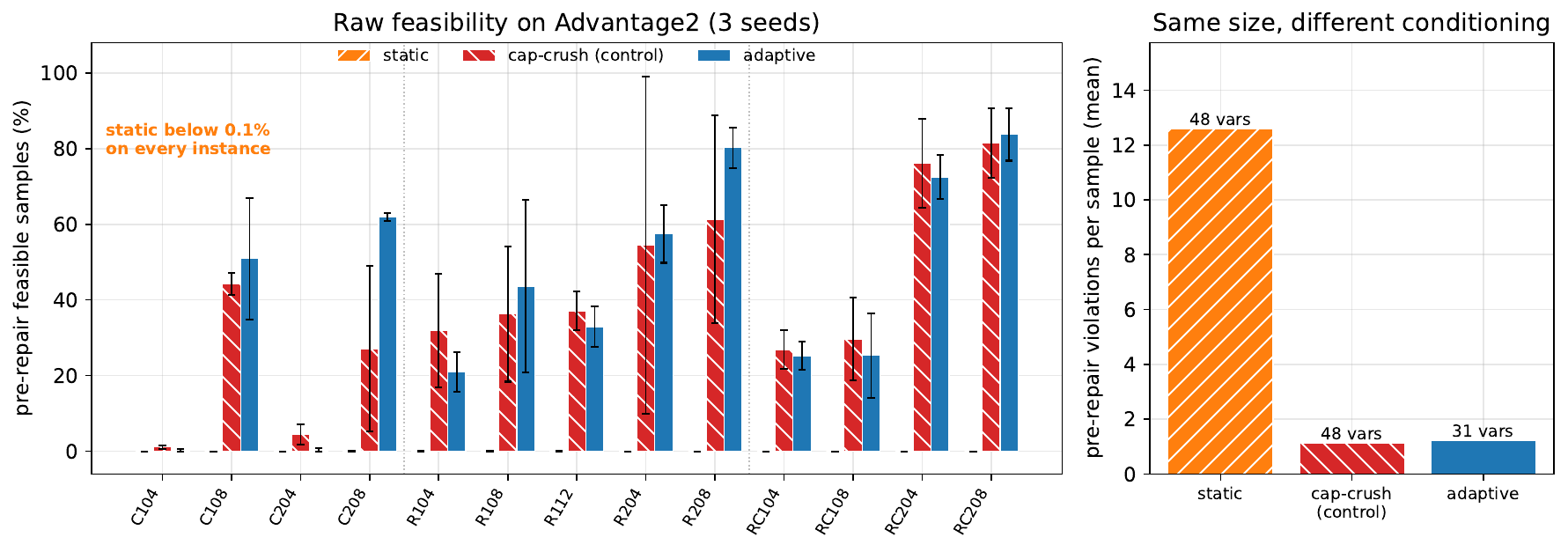}
\caption{Conditioning on Advantage2. Left: pre-repair feasible samples per instance
(mean $\pm$ s.d.\ over three seeds); static stays below $0.1\%$ on every instance.
Right: mean pre-repair violations with the logical variable count annotated. Static
and cap-crush use the same $48$ variables, so the gap between them isolates
dynamic-range conditioning from problem size.}
\label{fig:qpucond}
\end{figure}

The controlled comparison is between \emph{static} and \emph{cap-crush}. These
configurations have the same logical variable count ($47.8$ on average) and differ
in their coefficient conditioning. The static formulation produced no feasible
samples in $35$ of the $39$ runs and never exceeded $0.2\%$ feasibility in the
remaining four (Fig.~\ref{fig:qpucond}, Table~\ref{tab:qpu}). Cap-crush yielded
$39.4\%$ feasible samples and reduced the mean number of pre-repair violations from
$12.6$ to $1.1$ per sample. This fixed-variable comparison reproduces on hardware
the conditioning effect identified classically in Section~\ref{sec:phase2}. The
full adaptive formulation achieved a comparable feasibility rate ($42.7\%$) with
$35\%$ fewer variables.

\begin{table}[htbp]
\centering\small
\caption{D-Wave Advantage2, $13$ held-out instances $\times$ 3 seeds, $N{=}10$, GNN
coarsening, $1{,}000$ reads. Static and cap-crush share a variable count, so the
difference between them is conditioning alone. The Ising rescaling factor is the
factor by which the problem is compressed to fit the device h/J range.}
\label{tab:qpu}
\begin{tabular}{lcccc}
\toprule
penalties & pre-repair feas. & pre-repair viol. & QUBO vars & Ising rescaling \\ \midrule
static & 0.02\% & 12.6 & 48 & $2.7\times10^{10}$ \\
cap-crush (control) & 39.4\% & 1.13 & 48 & $1.7\times10^{5}$ \\
\textbf{adaptive} & \textbf{42.7\%} & \textbf{1.24} & \textbf{31} & $\mathbf{1.6\times10^{5}}$ \\
\bottomrule
\end{tabular}
\end{table}

\begin{figure}[htbp]
\centering
\includegraphics[width=\textwidth]{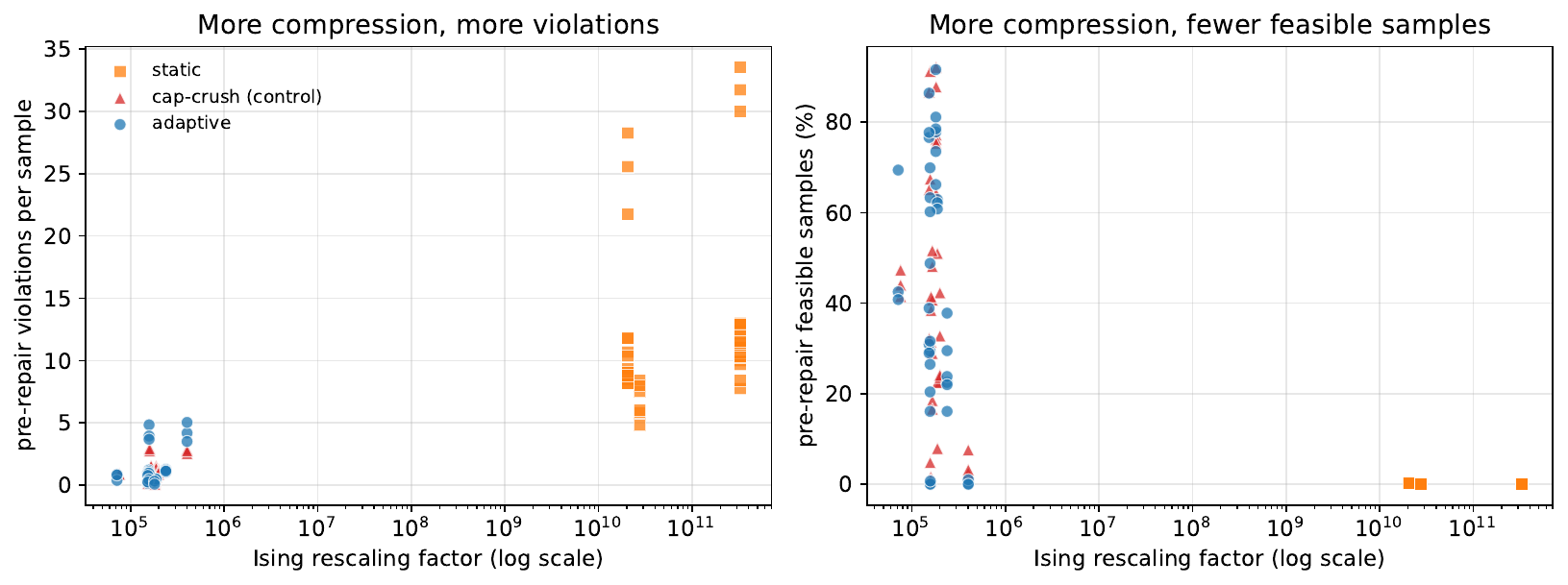}
\caption{The Ising rescaling factor, not the raw coefficient ratio, predicts raw
quality on hardware: the more the problem must be compressed to fit the device
ranges, the more of the objective falls below coupler resolution.}
\label{fig:qpumech}
\end{figure}

\textbf{The hardware-relevant measure of conditioning.} In
Section~\ref{sec:phase2}, we use the ratio of the largest to the smallest coefficient
as a conditioning proxy. On hardware, however, this ratio does not capture the
relevant loss of precision. Cap-crush has a higher median ratio than static
($3.3\times10^{4}$ versus $8.7\times10^{3}$), despite performing substantially
better, because reducing the capacity weight also reduces the smallest
coefficients. The physically relevant quantity is the factor required to rescale
the Ising problem into the device's $h$ and $J$ ranges. This compression determines
how much of the objective lies below the coupler resolution. The rescaling factor
differs by five orders of magnitude between the two arms
($2.7\times10^{10}$ for static and $1.7\times10^{5}$ for cap-crush) and varies
monotonically with raw solution quality (Fig.~\ref{fig:qpumech}).

\subsection{Scope of the hardware result}
Two limitations delimit the hardware result. First, the annealer does not
outperform the classical sampler on these instances. Simulated annealing reaches
$79\%$ pre-repair feasibility on the same coarsened QUBOs at $N{=}10$, compared
with $43\%$ on the processor, a difference consistent with the residual compression
required by the hardware. The experiment shows that conditioning affects whether
the processor can exploit a formulation; it does not establish an advantage over
classical sampling. Second, the accessible regime remains small
($N\approx10$--$20$; Section~\ref{sec:embed}). The hardware evaluation should
therefore be read as a conditioning study at presently embeddable scales, not as an
end-to-end routing benchmark.

\section{Beyond the position encoding}
\label{sec:encodings}
The conditioning burden identified above originates in the slack variables of the
position encoding. To examine whether it persists under other formulations, we
evaluate two slack-free encodings on the same instances and with the same
simulated-annealing budget.

\textbf{Route activation} assigns one binary variable to each directed arc rather
than to each route position. Degree, depot-balance, and short-cycle constraints are
imposed as penalties, while arcs that violate a time window individually are
excluded. The formulation is compatible with coarsening and reduces the QUBO size
by a factor of $4$--$7$ ($420\!\to\!59$--$75$ variables at $N{=}20$). This reduction
does not improve feasibility: at $N{=}20$, feasibility is $66.7\%$ without
coarsening and $55.6\%$ with GNN coarsening. Although the encoding removes locally
infeasible arcs, it does not represent cumulative travel time along a route.
Consequently, a sequence of individually admissible arcs may still violate a time
window. Route activation therefore provides an informative test of size reduction,
but not a complete feasibility formulation for CVRPTW.

\begin{figure}[htbp]
\centering
\includegraphics[width=0.62\textwidth]{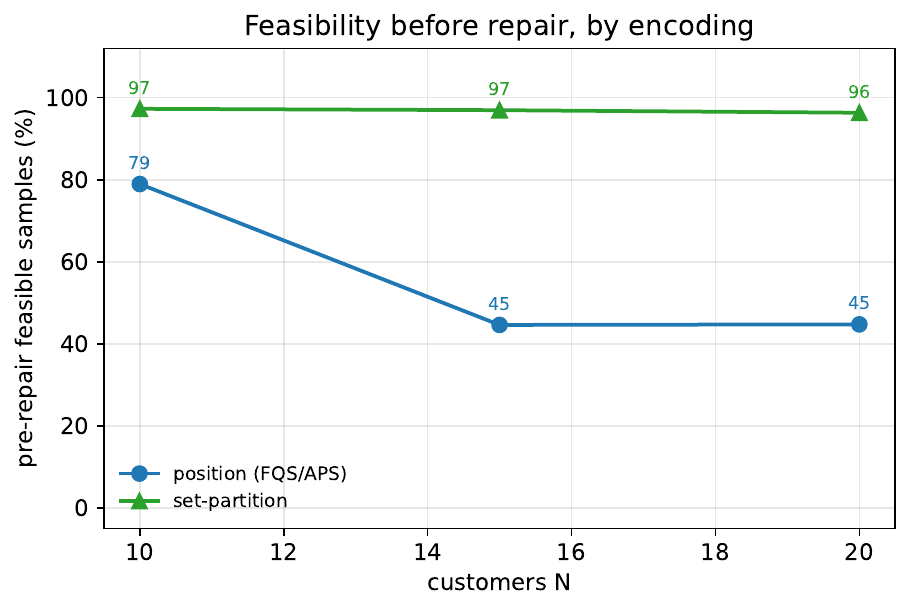}
\caption{Pre-repair feasibility by encoding (GNN coarsening, identical solver budget,
measured over the full sample distribution in both cases). Building feasibility into
the columns removes the slack-induced conditioning burden.}
\label{fig:encodings}
\end{figure}

\textbf{Set partitioning} incorporates feasibility at an earlier stage. A classical
procedure generates a pool of routes that already satisfy the capacity and
time-window constraints. Each binary variable activates one route, and the QUBO
selects a minimum-cost subset that covers every customer exactly once. Because
feasibility is encoded in the route columns, the formulation requires neither
capacity slack variables nor a one-hot position block. The raw samples improve
substantially (Fig.~\ref{fig:encodings}): exact-cover rates are $97\%$, $97\%$, and
$96\%$ at $N{=}10,15,20$, respectively, compared with pre-repair feasibility of
$79\%$, $45\%$, and $45\%$ for the position encoding. The corresponding route
pools contain only $56$--$90$ variables. This formulation has an important
structural qualification: the route pool is generated classically, leaving the
annealing backend to solve only the selection problem and placing it in direct
competition with classical set-partitioning methods. Nevertheless, the comparison
shows that the conditioning burden belongs to the encoding rather than inherently
to the routing problem. Set partitioning is therefore a promising direction for
hardware evaluation, but it is not yet an end-to-end alternative.

\FloatBarrier
\section{Discussion and limitations}
The most transferable finding concerns QUBO conditioning. For QUBOs with
slack-encoded constraints, increasing the absolute penalty scale has little effect
on the sampler, whereas controlling the range of coefficients improves the quality
of raw samples. The variable-count-preserving control separates this effect from a
reduction in problem size. On Advantage2, the same comparison increases raw
feasibility from $0.02\%$ to $39\%$ without changing the number of logical variables
(Section~\ref{sec:qpu}). The learned coarsener addresses a separate limitation of
the prior method by removing its dependence on family-specific hyperparameters.
Across $N{=}10$--$100$, it preserves feasibility more reliably than the tuned
heuristic, with a larger advantage at scales where the fixed heuristic parameters
transfer poorly. At the same time, coarsening keeps the QUBO solvable at $N{=}100$
and embeddable at $N{=}20$.

The study has five principal limitations. First, the hardware experiments
characterise the formulation, not the relative performance of the device. On the
same coarsened QUBOs, simulated annealing achieves higher pre-repair feasibility
than the quantum annealer. Moreover, minor embedding confines the hardware
evaluation to approximately $N{=}10$--$20$. These experiments consequently test
conditioning at the scale supported by the current device rather than benchmark an
end-to-end routing solver.

Second, the claims concern QUBO formulation and preprocessing, not final solution
cost. Classical repair followed by local search provides a reference bound that the
pipeline reaches but does not improve upon (Table~\ref{tab:phase4}). Adaptive
conditioning and feasibility-preserving coarsening are therefore evaluated using
the quantities they are designed to affect: raw sample quality, feasibility, and
tractability.

Third, coarsening reduces the problem size consistently by a factor of roughly five
to six. This is a substantial multiplicative reduction, but it does not change the
asymptotic growth of the underlying encoding. Choosing the coarsening depth
separately for each instance may extend the practical range further.

Fourth, the learned coarsener's feasibility advantage arises from its ability to
operate across families and sizes without retuning, rather than from the merge score
considered in isolation. Under matched settings, its difference in pre-repair
violations from a fixed-parameter heuristic is not significant ($p\!=\!0.62$ pooled
over $N$). Changing either the message-passing architecture or the reward
coefficients also has little effect on the final outcome
(Section~\ref{sec:robust}).

Finally, FQS and APS coincide under the configuration studied here, and all primary
experiments use the Solomon benchmark. Section~\ref{sec:encodings} provides an
initial test of formulation dependence. In particular, the set-partitioning result
suggests that the observed conditioning burden is specific to the position encoding
rather than intrinsic to CVRPTW.

\section{Conclusion and future work}
This work introduced adaptive penalty calibration and a learned coarsening policy
for a QUBO formulation of CVRPTW. Both components were evaluated with simulated
annealing and on a D-Wave Advantage2 processor. At a fixed sampling budget,
adaptive calibration reduces the mean number of raw constraint violations from
$33.0$ to $0.06$. A control that preserves the variable count attributes this
improvement to coefficient conditioning rather than to a smaller QUBO. The hardware
experiment yields the same distinction: the poorly conditioned formulation returns
almost no feasible samples, whereas the conditioned formulation reaches $39\%$ raw
feasibility with the same number of logical variables.

The learned coarsener uses one configuration across the Solomon families and
preserves feasibility more often than the family-tuned heuristic over
$N{=}10$--$100$, with the difference increasing at larger problem sizes. It also
reduces the QUBO size by a factor of approximately five to six, allowing the
formulation to remain solvable at $N{=}100$ and embeddable at $N{=}20$. These
results establish improvements in conditioning, feasibility, and tractability. They
do not establish an improvement in end-to-end solution cost: after classical repair
and local search, the pipeline matches but does not surpass the classical reference
bound.

The results motivate three directions for further work. First, the set-partitioning
encoding avoids the slack variables responsible for the conditioning burden and
achieves $96$--$97\%$ pre-repair feasibility. Evaluating this formulation on
hardware will require integrating route-pool generation into the optimization loop.
Second, reverse annealing could use the output of classical repair as an initial
state, shifting the role of the annealer from direct competition with the classical
procedure to local improvement. Third, a context-dependent stopping rule could
replace the fixed coarsening depth and address the loss of feasibility observed at
larger $N$.

\paragraph{Code and data availability.} All code, configurations, trained models,
and result files needed to reproduce every figure and table in this paper are
available at \url{https://github.com/YoussefKamelKamel1/cvrptw-qubo-coarsening}.
The Solomon benchmark instances are public and are included in the repository.

\paragraph{Acknowledgement}
This article was prepared as part of the project “Solving routing problems using quantum annealing”, mentored by Paweł Gora within the QIntern 2026 program organized by the QResearch Department of QWorld. The authors express sincere gratitude to QWorld and to the project team members for their valuable support and guidance throughout the development of this work.

We gratefully acknowledge the Polish high-performance computing infrastructure PLGrid (HPC Center: ACK Cyfronet AGH) for providing computer facilities and support within the computational
grant no. PLG/2026/019625.

\FloatBarrier
\small
\bibliographystyle{plainnat}
\bibliography{references}

\end{document}